\documentclass{article}

\usepackage{microtype}
\usepackage{graphicx}
\usepackage{subfigure}
\usepackage{booktabs} 

\usepackage{hyperref}

\usepackage[accepted]{icml2024}

\usepackage{amsmath}
\usepackage{amssymb}
\usepackage{mathtools}
\usepackage{amsthm}

\usepackage[capitalize,noabbrev]{cleveref}

\theoremstyle{plain}

\theoremstyle{definition}

\theoremstyle{remark}

\usepackage[textsize=tiny]{todonotes}
\usepackage{multirow}

\icmltitlerunning{MorphPiece : A Linguistic Tokenizer for Large Language Models}

\begin{document}

\twocolumn[
\icmltitle{MorphPiece : A Linguistic Tokenizer for Large Language Models}




\begin{icmlauthorlist}
\icmlauthor{Haris Jabbar}{yyy}
\end{icmlauthorlist}

\icmlaffiliation{yyy}{Ludwig Maximilian University, Munich, Germany}

\icmlcorrespondingauthor{Haris Jabbar}{haris.jabbar@gmail.com}

\icmlkeywords{Machine Learning, ICML, NLP, LLM, Large Language Models, Tokenization}

\vskip 0.3in
]



\printAffiliationsAndNotice{}  

\begin{abstract}
Tokenization is a critical part of modern NLP pipelines. However, contemporary tokenizers for Large Language Models are based on statistical analysis of text corpora, without much consideration to the linguistic features. I propose a linguistically motivated tokenization scheme, MorphPiece, which is based partly on morphological segmentation of the underlying text. A GPT-style causal language model trained on this tokenizer (called MorphGPT) shows comparable or superior performance on a variety of supervised and unsupervised NLP tasks, compared to the OpenAI GPT-2 model. Specifically I evaluated MorphGPT on language modeling tasks, zero-shot performance on GLUE Benchmark with various prompt templates, massive text embedding benchmark (MTEB) for supervised and unsupervised performance, and lastly with another morphological tokenization scheme (FLOTA \cite{flota}) and find that the model trained on MorphPiece outperforms GPT-2 on most evaluations, at times with considerable margin, despite being trained for about half the training iterations.
\end{abstract}

\section{Introduction}

Modern NLP pipelines involve segmenting text into discrete units which are represented with learnable high dimensional vectors. This segmentation, called tokenization, forms the basis of most transformer (and pre-transformer e.g. LSTM, RNN \cite{pennington2014glove, mikolov, bojanowski, elmo}) based architectures. Many tokenization algorithms have been explored over the past few years, ranging from characters to words and an intermediate form called subword tokenization. The most commonly used tokenizers such as BPE \cite{bpe}, WordPiece \cite{wordpiece}, Unigram \cite{unigram} etc) follow the subword tokenization paradigm, which relies on the statistical properties of the corpora to construct the tokenization algorithm. The statistical properties used by these tokenizers usually involve finding sub-words that are either most frequent (BPE, WordPiece etc.) or ones that optimize some loss (Unigram). Arguably, these schemes do not consider any linguistic information and instead deal with language at purely symbolic level. So for example similar techniques can and have been applied in other symbolic domains, e.g. genome sequencing \cite{dnabert,genome}. However, natural language is far more than just a collection of symbols. From morphology to phonetics; from syntax to phonology; there is a plethora of information that goes unutilized when we treat language as mere symbols. In this work, I make an attempt to change that by incorporating language morphology during the tokenization process.

\begin{table} 
\centering
\begin{tabular}{cccc}
\toprule
\textbf{Task} & \textbf{GPT-2} & \textbf{MorphGPT} & \textbf{Templates} \\ \midrule
\textbf{RTE} & 0.216 & \textbf{0.335 \small{(+54\%)}} & \textbf{5/6} \\
\textbf{SST} & \textbf{0.468} & 0.466 & 4/8 \\
\textbf{QQP} & 0.183 & \textbf{0.235\small{(+28\%)}} & \textbf{4/5} \\
\textbf{WNLI} & \textbf{0.386} & 0.35 & 1/5 \\
\textbf{MRPC} & 0.666 & \textbf{0.673} & 2/5 \\
\textbf{COLA} & 0.622 & \textbf{0.655} & 2/5 \\
\textbf{QNLI} & 0.144 & \textbf{0.211\small{(+46\%)}} & \textbf{3/5} \\
\textbf{MNLI} & 0.241 & \textbf{0.288\small{(+19\%)}} & \textbf{8/10} \\ \bottomrule
\end{tabular}
\caption{Zero Shot evaluation of GLUE benchmark with multiple prompts per task using PromptSource \cite{promptsource}. Last column shows ratio of templates where MorphGPT outperforms GPT-2}
\label{tab:glue_zeroshot}
\end{table}

It has been previously shown \cite{dagobert, superbizarre} that morphologically informed vocabularies lead to better generalization capabilities of language models. In this work, I build on this insight and propose a tokenization approach that relies partly on the morphological construction of words to break them down into sub-words. The intuition being that sub-words so constructed will be more natural splits than a split on statistical properties, and hence might lead to more efficient models. For instance, "paratrooper" would be segmented as (’para\#’, ’troop’, ’\#er’) in our tokenizer, which aligns more closely with the linguistic parts of the word compared to the BPE and Wordpiece tokenizers that split it into (’par’, ’atro’, ’oper’) and ('para', '\#\#tro', '\#\#oper'), respectively. In the proposed tokenizer, I combine a deterministic morphological segmentation algorithm with the statistical BPE algorithm. To validate the approach, I train a GPT-like architecture with the proposed tokenizer and compare it with the OpenAI GPT-2 model that uses BPE tokenization. The results demonstrate that this tokenizer leads to improved performance across a wide range of NLP tasks. \\

\begin{table*} 
\centering
\begin{tabular}{@{}cccc@{}}
\toprule
\textbf{Word} & \textbf{\begin{tabular}[c]{@{}c@{}}BPE\\ tokens\end{tabular}} & \textbf{\begin{tabular}[c]{@{}c@{}}Wordpiece\\ tokens\end{tabular}} & \textbf{\begin{tabular}[c]{@{}c@{}}MorphPiece\\ tokens\end{tabular}} \\ \midrule
batting & 'bat', 'ting' & batting & 'bat', '\#ing' \\
disengage & 'dis', 'eng', 'age' & 'di', '\#\#sen', '\#\#ga', '\#\#ge' & 'dis\#', 'en\#', 'gage' \\
archeologists & 'ar', 'che', 'ologists' & 'arch', '\#\#eo', '\#\#logists' & 'archaeo\#', '\#logy', '\#ist', '\#s' \\
eyewitnesses & 'ey', 'ewitness', 'es' & 'eye', '\#\#wi', '\#\#tness', '\#\#es' & 'eye', '\#', 'wit', '\#ness', '\#s' \\
photographers & 'phot', 'ographers' & photographers & 'photo\#', '\#graph', '\#er', '\#s' \\ \bottomrule
\end{tabular}
\caption{Comparison of tokens produced by different tokenization schemes}
\label{tab:tokenization_comparison}
\end{table*}

The tokenizer is called MorphPiece\footnote{There is a similarly named R library (morphemepiece) : https://github.com/macmillancontentscience/morphemepiece} and  
Table \ref{tab:tokenization_comparison} gives some examples that highlight the manner in which MorphPiece splits words compared to BPE and Wordpiece. Few aspects are apparent here :

\begin{enumerate}
    \item The location and presence/absence of \# symbol denotes prefixes, suffixes, compound words and stems. In particular \# at the end (resp. beginning) of a subword is a prefix (resp. suffix), a \# as a standalone subword denotes compound words and a subword without a \# denotes the word stem. 
    \item MorphPiece segmentation splits up the words into liguistically aligned affixes which have a semantic meaning. This is not the case with statistical tokenizers.
    \item MorphPiece uses canonical (instead of surface form) segmentation by modifying the spellings without which this alignment would not be possible (e.g. batting is tokenized as ['bat','ing'], instead of ['bat','ting'] in BPE).
    \item Such splitting into affixes opens up potential analyses of suffix and prefix representations, that isn't possible with statistical tokenizers. For example, unlike BPE/WordPiece, the negation prefixes like 'de' and 'un' and 'dis' are clearly segmented from the stem in MorphPiece.

\end{enumerate}

The primary contributions of this paper are as follows :

\begin{enumerate} 
    \item I propose a linguistically motivated tokenizer that results in superior performance across a wide variety of NLP tasks, compared to the same architecture trained on BPE.
    \item I also propose an algorithm for combining the tokens back into words and sentences,
    \item Upon publication of the paper, I will open-source the code and various checkpoints of the MorphGPT model trained on MorphPiece.
\end{enumerate}

\begin{figure*}[!htb]
    \includegraphics[width=0.9\textwidth]{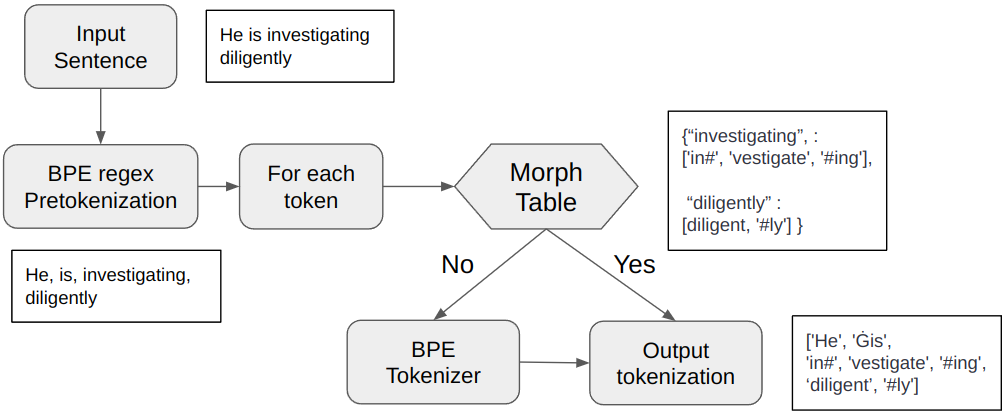}
    \centering
    \caption{MorphPiece tokenization Scheme : After standard BPE pre-tokenization, for each pre-token, we tokenize with MorphTable if the token exists in the MorphTable; if not, we apply standard BPE with custom trained vocabulary}
    \label{fig:scheme}
\end{figure*}

\section{Related Work}
\label{relatedwork}

There is ample body of research for building morphological tokenizers using either supervised, unsupervised or manual curation methods. Morfessor \cite{morfessor} and its variants \cite{morfessor2,gronroos-etal-2020-morfessor} are the most well known. In SIGMORPHON 2022 Shared Task for Morpheme Segmentation \cite{batsuren-etal-2022-sigmorphon}, there were 13 submissions to build morpheme segmentation at word and sentence level. This challenge was itself built on another Morpho-Challenge series \cite{morphochallenge}. However, in all these challenges, unlike this work, achieving a morphological segmentation was the end goal. Whereas MorphPiece takes morphological segmentation as the starting point and builds a language model with it.\\

Use of morphological segmentation for tokenization has been explored extensively in the context of Neural Machine Translation with mixed results \cite{pan,zhou,domingo,machacek,saleva-lignos,banerjee,ataman-federico}. However, use of morphological analysis in language modeling, especially on transformer-based architectures, is rather limited. \citet{bpe_suboptimal} compare BPE and Unigram tokenization for morphological alignment and find that Unigram is more aligned to morphpological splits, and leads to better or similar performance in downstream tasks. Similarly \citet{dagobert,superbizarre} showed that a morphologically informed vocabulary improves performance of LLMs. Subsequently \citet{flota} proposed a statistical tokenization improvement method (FLOTA) that tries to align the tokenization with morphological segmentation and show that this improves the performance on a specific task. This work is different from FLOTA in a couple of important ways. First, unlike MorphPiece, they use statistically built vocabulary of WordPiece/BPE/Unigram. Second, they apply their method only during fine-tuning stage, unlike this work where a model is pre-trained from scratch. Third, they don't have separate deterministic (morphological) and statistical modes of tokenization. Fourth, they evaluate on only one custom built task. Finally, MorphGPT model outperforms FLOTA on that task, by a considerable margin. (Section \ref{flota}).

In languages other than English, especially those which are considered morphologically rich, there are few models e.g Hebrew \cite{alephbert}, Arabic \cite{arab,arabert}, Turkish \cite{turkish} and Russian/Finnish \cite{matthews}. This work differs from these, in two important ways. First, it is applied to English which is not known to be a morphologically rich language and second, we use a causal language model, whereas the quoted works have all used BERT-based masked language models. 

\section{MorphPiece}
\label{morphpiece}

In this section, I present MorphPiece; an English language tokenization scheme that combines Byte Pair Encoding (BPE) with morpheme based segmentation for a more linguistically aligned tokenization mechanism. The tokenization scheme is shown in Figure \ref{fig:scheme}. First, the text is normalized and pre-tokenized as per the standard BPE tokenization \cite{bpe}. In case of BPE, these pretokens are a regex based splitting of sentences. These pretokens are then passed through a look-up table of words (called MorphTable), to see if a morpheme based segmentation is available. If a segmentation is found, the pretoken is replaced with the corresponding morphemes; if not, the tokens are split according to the BPE tokenization scheme with a custom-trained vocabulary. 

\subsection{MorphTable}

MorphTable is a dictionary with keys being the words from English language and values being their respective morphological segmentation. To construct MorphTable, I used MorphyNet \cite{morphynet}, which is a database of derivational and inflectional morphology for 15 languages. I constructed a lookup table of 346,340 words from English language which have been segmented into morphemes from the database. Table \ref{table:morph_freq} shows the frequency count of these segmentations. The extreme high number of morphemes comes from chemical compounds (e.g. dichlorodiphenyltrichloroethane). For the purpose of MorphPiece, I created a vocabulary from the set of unique affixes and stems from MorphTable after dropping the entities with fewer than 5 occurrences on our training corpus. This trimmed version had 18,304 tokens, and table size of 134,943 entries.

\subsection{MorphPiece Vocabulary}

The MorphPiece vocabulary has two sources. First, we have the MorphTable described above. All the affixes and stems from this table are added to the vocabulary. The second component is the trainable BPE vocabulary. In the spirit of fair comparison, I aimed for the same vocabulary size as that of GPT-2, i.e 50,257 tokens. Accounting for the vocabulary from MorphTable (18,304 tokens), I trained a BPE tokenizer to build a vocabulary size of 32,000. I used OpenWebText \cite{owt} as the training corpus. Before training this tokenizer, I removed all words that had a segmentation available in the MorphTable: the idea being that since those words will be processed by the MorphTable and not by the BPE algorithm, hence the BPE tokenizer should be trained only on the text that will be tokenized by it. After merging the two vocabularies and accounting for a few common tokens, the final vocabulary size is 50,006.

\subsection{MorphPiece Fertility} 
Fertility is defined as the average number of subwords that a tokenizer splits a word into. So the tokenization ('para', '\#\#tro', '\#\#oper') of the word 'paratrooper' has a fertility of 3. When averaged over a large corpus, fertility is a measure of how aggressively a tokenizer splits the words. This has significance for perplexity comparison. To measure fertility, I tokenized the training dataset with both BPE and MorphPiece, and additionally with a whitespace splitter as a proxy for number of words in a sentence. Subsequently, I compared the average number of tokens produced by the tokenizers for various sentence lengths. MorphPiece produced about 17\% longer sentences compared to BPE. This may be attributed to reduced BPE vocabulary available to MorphPiece, which results in more aggressive tokenization. 

\section{The Language Model : MorphGPT}
\label{morphgpt}
A concrete test of any new tokenization scheme is from the performance of a language model trained on that scheme, on various NLP tasks. Towards that end, I trained a GPT-2 (Base) architecture with MorphPiece and compared it with the OpenAI GPT-2 model which uses BPE. The model is called MorphGPT-Base. It is pertinent to note that, other than the tokenization scheme, MorphGPT-Base does not have any architectural differences from GPT-2 (Base).

\subsection{Training setup}
GPT-2 was trained on a custom built corpus called WebText. Since that corpus is not available publicly, I used its open source clone, called the OpenWebText \cite{owt}. Additionally, I used HuggingFace's implementation of GPT-2 \cite{huggingface} with Pytorch-Lightning \cite{pytorch} as the training framework on Nvidia A-100 GPUs. 

\begin{table}[ht]
    \centering
    \begin{tabular}{@{}ccc@{}}
        \toprule
        \textbf{\begin{tabular}[c]{@{}c@{}}Number of \\ Morphemes\end{tabular}} & {\bfseries MorphyNet} & {\bfseries MorphPiece} \\
        \midrule
        2 & 136,715 & 67,169 \\
        3 & 143,990 & 48,264 \\
        4 & 54,129 & 16,670 \\
        5 & 10,001 & 2,589 \\
        6 & 1,236 & 217 \\
        7 & 208 & 24 \\
        8+ & 61 & 10\\
        \midrule
        Total & 346,340 & 134,943 \\
        \bottomrule
    \end{tabular}
    \caption{Frequency of Morpheme Segmentations from MorphyNet \cite{morphynet} and MorphPiece}
    \label{table:morph_freq}
\end{table}

I trained the MorphGPT model for a total of 200k steps and compared its performance with the GPT-2 (Base) model available on HuggingFace hub as 'gpt2' checkpoint. For training hyperparameters, it is pertinent to note that except for batch size of 512, original GPT-2 parameters are not available. To estimate training iterations, I used the GPT-2 models trained by the Stanford CRFM’s Mistral team\footnote{https://github.com/stanford-crfm/mistral}. I evaluated perplexity scores for Lambada, PTB and Wikitext103 for the checkpoints from 50k to 400k and compared the values with GPT-2 perplexity scores. With this proxy, I estimated that GPT-2 was trained for about 400k-500k steps (Table \ref{tab:mistral_gpt2}).

\begin{table}
\centering
\begin{tabular}{@{}cccc@{}}
\toprule
\textbf{\begin{tabular}[c]{@{}c@{}}Mistral\\ GPT-2 (Small)\end{tabular}} & \textbf{PTB} & \textbf{WikiText} & \textbf{Lambada} \\ \midrule
\textbf{50k} & 78.4239 & 40.0234 & 72.7688 \\
\textbf{100k} & 72.935 & 37.4006 & 71.9893 \\
\textbf{200k} & 66.1411 & 34.889 & 66.4272 \\
\textbf{300k} & 64.8948 & 33.5991 & 62.4198 \\
\textbf{400k} & 59.949 & 31.4337 & 60.1416 \\ \midrule
\textbf{GPT-2} & 60.9468 & 29.769 & 56.0281 \\ \bottomrule
\end{tabular}
\caption{Perplexity scores of OpenAI GPT-2 (Base) with Mistral GPT-2 (Small) models at various training checkpoints. It shows that GPT-2 was likely trained for around 400k steps.}
\label{tab:mistral_gpt2}
\end{table}

Additionally I used batch size of 512 and one-cycle learning rate scheduler \cite{onecycle} with maximum learning rate of $1e^{-3}$, warm-up of 2000 steps, and cosine decay to final learning rate of $1e^{-5}$. For the optimizer, I used Adam \cite{adam} with betas 0.9 and 0.995 and eps of $1e^{-8}$.

\section{Evaluations}
\label{evaluations}

MorphGPT was evaluated on a wide variety of tasks. Specifically evaluations were conducted on tasks closely related to language modeling (perplexities on various datasets, and LAMBADA task); seven different sequence embedding evaluations tasks (both supervised and unsupervised) using Massive Text Embedding Benchmark (MTEB) and zero shot prompt-based evaluations on GLUE. In almost all categories MorphGPT shows superior performance across the board, despite being trained for almost half the steps. Finally, I compared MorphGPT with a model trained on similarly themed tokenization scheme called FLOTA \cite{flota} and find that MorphGPT also outperforms this method comprehensively.

\subsection{Language Modeling}
I evaluated MorphGPT-Base and GPT-2 (Base and Large) on Penn Tree Bank \cite{ptb}, OpenAI GPT-Outout-Dataset\footnote{https://github.com/openai/gpt-2-output-dataset} and LAMBADA datasets \cite{lambada}. As can be seen in Table \ref{tab:perplexity}, MorphGPT models show much better token-level perplexity numbers over fully trained GPT-2 models. In particular, even with only 50k steps, MorphGPT achieves better perplexity than GPT-2 (Base) across all three datasets; and reaches performance of GPT-2 (Large) with 200k steps. Generally token level perplexity is not comparable across different vocabularies \cite{cotterell}. However. as we will see below, MorphGPT not only shows better performance on a wide variety of downstream tasks, but also on LAMBADA, which is closely related to language modeling task. Additionally, Section \ref{discussion} has more discussion on this issue.

\paragraph{LAMBADA} To reconfirm performance on language modeling, I evaluated a cloze style LAMBADA \cite{lambada} task. The task is to predict the last word of a paragraph and it is designed in a way that local context is not enough and one requires the whole paragraph to predict the correct answer. This task is known to be particularly hard for models to do well on \cite{gpt3}. MorphGPT surpasses the accuracy of GPT-2 by almost 10\% with only 50k steps and almost reaches the accuracy of a six-times larger GPT-2 Large model (Table \ref{tab:perplexity}). This confirms that token-level perplexity gains translate to better performance in other related tasks.

\begin{table}
\centering
\begin{tabular}{@{}l|cc@{}}
\toprule
\textbf{LAMBADA} & \textbf{GPT-2} & \textbf{MorphGPT} \\ \midrule
With Fragments & 46.88 & \textbf{58.58 \small{(+25\%)}} \\
Without Fragments & 32.5 & \textbf{36.6 \small{(+13\%)}} \\
Fragment Ratio & 0.778 & 0.644 \\ \bottomrule
\end{tabular}
\caption{Accuracy of MorphGPT and GPT-2 on LAMBADA last word prediction task, when the word is fragmented/intact after tokenization.}
\label{tab:lambada}
\end{table}

Additionally, to cater for difference in tokenizations, I evaluated a subset of LAMBADA in which the word to be predicted is not fragmented by tokenization (Table \ref{tab:lambada}). Even in this configuration, MorphGPT shows double digit improvement over GPT-2. 

\begin{table*}
\centering
\begin{tabular}{@{}lccccccc@{}}
\toprule
\textbf{} & \textbf{Metric} & \textbf{\begin{tabular}[c]{@{}c@{}}GPT-2\\ Base\end{tabular}} & \textbf{Morph50} & \textbf{Morph100} & \textbf{Morph150} & \textbf{Morph200} & \textbf{\begin{tabular}[c]{@{}c@{}}GPT-2\\ Large\end{tabular}} \\ \midrule
\textbf{PennTreeBank} & ppl & 61.859 & \textbf{43.692} & \textbf{39.841} & \textbf{38.691} & \textbf{38.251} & 37.931 \\
\textbf{OpenAI-250k} & ppl & 25.544 & \textbf{18.743} & \textbf{17.893} & \textbf{17.505} & \textbf{17.262} & 16.743 \\
\textbf{Lambada} & ppl & 55.868 & \textbf{47.423} & \textbf{45.398} & \textbf{43.258} & \textbf{42.846} & 37.211 \\ 
\textbf{Lambada} & acc & 46.88 & \textbf{55.19} & \textbf{56.57} & \textbf{58.21} & \textbf{58.5} & 58.76 \\\bottomrule
\end{tabular}
\caption{Evaluation of various datasets on token level perplexity and LAMBADA task}
\label{tab:perplexity}
\end{table*}

\subsection{Zero Shot evaluation on GLUE}

I evaluated MorphGPT on GLUE \cite{glue} in a zero shot setting with no in-context learning. Essentially, the evaluation consisted of turning each sample into a prompt and evaluating it on an appropriate metric. Additionally the labels were picked in a way that tokenization doesn't split it into subwords. Since prompt based evaluations are known to depict high variability depending on the wording of prompting template \cite{koksal}, I used multiple templates from PromptSource \cite{promptsource}. The results (Table 4) show that (except for WNLI), MorphGPT shows either comparable or much better performance, both in raw accuracy percentage and number of prompt templates where it outperforms GPT-2. 

\subsection{Sequence Embedding}

To test the performance of MorphGPT sequence embeddings, I evaluated it on Massive Text Embedding Benchmark (MTEB) \cite{mteb} which consists of 8 embedding tasks (both supervised and unsupervised) covering a total of 58 datasets. MorphGPT outperforms GPT-2 across all 7 monolingual tasks (omitted multilingual Bitext Mining task), often with considerable margin (Table \ref{tab:mteb}). To construct sequence embeddings, I took the average across tokens, of the last hidden state before softmax. For evaluations, I used the github repo from MTEB authors\footnote{https://github.com/embeddings-benchmark/mteb}.

\begin{table}
\centering
\begin{tabular}{@{}l|c|ccc@{}}
\toprule
\multicolumn{1}{c|}{\textbf{Task}} & \textbf{Metric} & \textbf{GPT-2} & \textbf{\begin{tabular}[c]{@{}c@{}}Morph\\ GPT\end{tabular}} & \textbf{Diff (\%)} \\ \midrule
\textbf{Clustering} & V\_Meas & 0.124 & \textbf{0.239} & 92.7 \\ \midrule
\textbf{Classification} & Accuracy & 0.459 & \textbf{0.537} & 17 \\ \midrule
\textbf{Summarization} & Spearman & 0.286 & \textbf{0.295} & 3.1 \\ \midrule
\textbf{Reranking} & MRR & 0.4 & \textbf{0.447} & 11.7 \\
\textbf{} & MAP & 0.322 & \textbf{0.361} & 12.1 \\ \midrule
\textbf{\begin{tabular}[c]{@{}l@{}}Pair \\ Classification\end{tabular}} & AP & 0.264 & \textbf{0.422} & 59.8 \\
\textbf{} & F1 & 0.291 & \textbf{0.431} & 48.1 \\ \midrule
\textbf{Retrieval} & \begin{tabular}[c]{@{}c@{}}Recall\\ @100\end{tabular} & 0.051 & \textbf{0.122} & 139.2 \\
\textbf{} & \begin{tabular}[c]{@{}c@{}}MAP\\ @100\end{tabular} & 0.011 & \textbf{0.028} & 54.5 \\ \midrule
\textbf{STS Tasks} & Spearman & 0.317 & \textbf{0.468} & 47.6 \\
\textbf{} & Pearson & 0.217 & \textbf{0.440} & 102.7 \\ \bottomrule
\end{tabular}
\caption{Results on Massive Text Embedding Benchmark (MTEB)}
\label{tab:mteb}
\end{table}

\paragraph{Classification} It consists of 12 datasets including intent, review, polarity and other classification tasks. The datasets embeddings are trained with logistic regression for 100 iterations and and evaluated on a test dataset. Results are reported as accuracy. MorphGPT outperforms GPT-2 in 11 out of 12 datasets  (Table \ref{tab:mteb_classification} in Appendix).

\paragraph{Clustering} It consists of 11 datasets of sentences or paragraphs and the goal is to cluster their embeddings in meaningful groups using a mini-batch k-means model. The clusters are measured using V-Measure \cite{v-measure}. MorphGPT outperforms GPT-2 in all 11 datasets with better scores and lower variance (Table \ref{tab:mteb_clustering}).

\paragraph{Pair Classification} It consists of 3 datasets of pairs of text inputs which need to classified as duplicate or otherwise. Various distance metrics (cosine similarity, dot product, euclidean distance, manhattan distance and maximum) between the embeddings of each pair are calculated and using best binary threshold, F1 and average precision are reported. Here again, MorphGPT outperforms GPT-2 across all distance metrics with more than 100\% improvement across first three metrics. (Table \ref{tab:mteb_pair} in Appendix).

\paragraph{Retrieval} This task consists of 11 datasets consisting of queries, a corpus of documents and a mapping from queries to relevant document in the corpus. The aim is to retrieve relevant documents for each query. After embedding the queries and documents with either MorphGPT or GPT-2, similarity scores are computed using cosine similarity and documents are ranked based on these scores. Since some of the datasets are huge, I evaluated on a subset of the 11 datasets, namely : ArguAna, Scidocs, Touche2020, TRECCOVID and 12 CQADUPStack datasets. MorphGPT outperforms GPT-2 across all datasets (Table \ref{tab:mteb_retrieval} in Appendix).

\paragraph{Semantic Textual Similarity (STS)} This task consists of 10 datasets of pair of texts with an associated label of how similar are the texts in a pair. Distance between pair embeddings from MorphGPT/GPT-2 are evaluated using various distance metrics and ranked. They are then compared with ground truth rankings and pearson/spearman correlations are reported. MorphGPT is better than GPT-2 by 50\% on spearman and 112\% on pearson correlation with cosine similarity.  (Table \ref{tab:mteb_sts} in Appendix).

\paragraph{Summary Evaluation} There is one dataset in this task; which consists of a set of human written and machine generated summaries. For each machine generated summary, the closest human summary is selected based on cosine similarity (or dot product) of embeddings. Subsequently, pearson/spearman correlations are computed with ground truth labels. Here MorphGPT performs only slightly better then GPT-2 on cosine similarity, but shows double digit improvement when using dot product. (Table \ref{tab:mteb_summary} in Appendix).

\paragraph{Re-ranking} In this task, there is a list of queries with an associated collection of documents with varying degree of relevance. The task is to rerank the documents according to relevance to each query, using the sequence embeddings from a model (MorphGPT or GPT-2) and cosine similarity as a distance metric. The final score is MAP/MRR of all queries. In this task we have 4 datasets and MorphGPT is better than GPT-2 in 3 of them, with an average improvement of about 15\%.  (Table \ref{tab:mteb_reranking} in Appendix).

\begin{table}
\centering
\begin{tabular}{@{}ccccc@{}}
\toprule
\textbf{}                              & \multicolumn{2}{c}{\textbf{ArXiv-L}}                 & \multicolumn{2}{c}{\textbf{ArXiv-S}} \\ \midrule
\multicolumn{1}{c|}{\textbf{Model}}    & \textbf{Dev}   & \multicolumn{1}{c|}{\textbf{Test}}  & \textbf{Dev}      & \textbf{Test}    \\ \midrule
\multicolumn{1}{c|}{\textbf{GPT-2}}    & 0.527          & \multicolumn{1}{c|}{0.51}           & 0.326             & 0.347            \\
\multicolumn{1}{c|}{\textbf{+FLOTA}}   & 0.553          & \multicolumn{1}{c|}{0.536}          & 0.35              & 0.371            \\
\multicolumn{1}{c|}{\textbf{MorphGPT}} & \textbf{0.667} & \multicolumn{1}{c|}{\textbf{0.652}} & \textbf{0.511}    & \textbf{0.523}   \\ \midrule
\multicolumn{1}{c|}{\textbf{Diff (\%)}} & \textbf{26.56} & \multicolumn{1}{c|}{\textbf{27.84}} & \textbf{56.75}    & \textbf{50.72}   \\ \bottomrule
\end{tabular}
\caption{Accuracy comparison of MorphGPT with GPT-2 and GPT-2+FLOTA (without noise). Diff shows performance improvement over GPT-2.}
\label{tab:flota_perf_val}
\end{table}

\begin{table}
\centering
\begin{tabular}{ccccc}
\toprule
\textbf{}                              & \multicolumn{2}{c}{\textbf{ArXiv-L (N)}}             & \multicolumn{2}{c}{\textbf{ArXiv-S (N)}} \\ \midrule
\multicolumn{1}{c|}{\textbf{Model}}    & \textbf{Dev}   & \multicolumn{1}{c|}{\textbf{Test}}  & \textbf{Dev}        & \textbf{Test}      \\ \midrule
\multicolumn{1}{c|}{\textbf{GPT-2}}    & 0.418          & \multicolumn{1}{c|}{0.406}          & 0.245               & 0.277              \\
\multicolumn{1}{c|}{\textbf{+FLOTA}}   & 0.46           & \multicolumn{1}{c|}{0.445}          & 0.25                & 0.266              \\
\multicolumn{1}{c|}{\textbf{MorphGPT}} & \textbf{0.586} & \multicolumn{1}{c|}{\textbf{0.568}} & \textbf{0.462}      & \textbf{0.463}     \\ \midrule
\multicolumn{1}{c|}{\textbf{Diff (\%)}} & \textbf{40.19} & \multicolumn{1}{c|}{\textbf{39.9}} & \textbf{88.57}    & \textbf{67.14}   \\ \bottomrule
\end{tabular}
\caption{Accuracy comparison of MorphGPT with GPT-2 and GPT-2+FLOTA (with noise).Diff shows performance improvement over GPT-2.}
\label{tab:flota_noise_perf_val}
\end{table}

\subsection{FLOTA}
\label{flota}

Finally, I present a comparison baseline. Few Longest Token Approximation (FLOTA) \cite{flota} is a tokenization improvement method that uses the vocabulary of standard BPE tokenizer but tries to preserve the morphological structure of words during tokenization. It achieves that by finding a segmentation that recursively finds the largest segment of a word and splits on that segment. For example the word 'undesirable' would be split as ('und', 'es', 'irable') by BPE; but with FLOTA, it will be split as ('un','desirable'); which is closer to exact morphological segmentation of  ('un','desire','able') used by MorphPiece. The authors show that FLOTA scheme preserves morphological structure to a large extent and that such a mechanism improves upon vanilla GPT-2 model. MorphPiece is different from FLOTA in a couple of important ways (Please see Section \ref{relatedwork} for details), however since this technique is closest to this work, we look at it in detail.

FLOTA was evaluated on a classification task of a custom dataset consisting of titles from computer science, maths and physics domains of ArXiv. A small (2000 samples) and a large (20,000 samples) dataset was constructed for each of the three areas. The models were finetuned for 20 epochs and evaluated on the dev/test splits. We adopted the exact same methodology and the results (Table \ref{tab:flota_perf_val}) show MorphGPT outperforms FLOTA comprehensively. While GPT-2+FLOTA shows an improvement of about 6\% over vanilla GPT-2 across both datasets, MorphGPT shows improvements of more than 35\%. Additionally the authors of FLOTA injected noise during evaluation to test the robustness of their scheme (Table \ref{tab:flota_noise_perf_val}). Here also, MorphGPT shows marked improvement over vanilla GPT-2 (40 \% in ArXiv-Large and 77 \% on ArXiv-Small). 
For more fine grained results, please refer to Tables \ref{tab:flota_all} and \ref{tab:flota_noise_all} and Figure \ref{fig:flota} in the Appendix.

\section{Detokenization}
\label{detokenization}
We define detokenization as the process of combining individual tokens produced by a model trained with MorphPiece (e.g MorphGPT), to form a sentence. While detokenization is straightforward for BPE and other statistical tokenizers, this is not the case with MorphPiece. This is primarily due to the fact that in MorphPiece, tokens come from one of two different sources: MorphTable or internal-BPE. During detokenization, we need to not only ascertain the source of individual tokens, but also find word boundaries and how to combine together the morphemes back to English words. These steps are describe in following sections.

\begin{figure*}
    \includegraphics[width=0.75\textwidth]{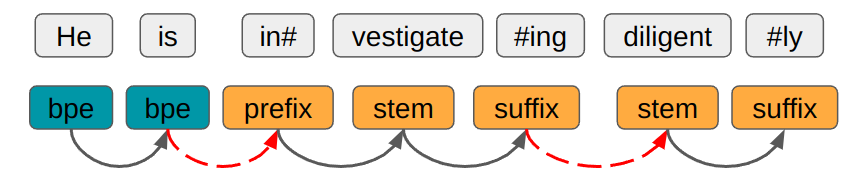}
    \centering
    \caption{Example of detokenization. First we mark the types of affixes or 'bpe'. Orange color tokens come from MorphPiece and teal colored come from BPE. Then we follow the arrows from Figure \ref{fig:detokenizer} to find word boundaries, which are looked up in reverse-MorphTable to find words.}
    \label{fig:detoken_example}
\end{figure*}

\subsection{Classification of Tokens}
\label{classify_tokens}
In the first stage, I use the token surface forms to classify the tokens as either 'morph' or 'bpe'; signifying the source they come from. Additionally the 'morph' tokens are annotated as either prefix, suffix, stem or hash (for compound words). MorphPiece tokens have four different surface forms. (a) The prefixes and suffixes have a '\#' sign at the end or beginning of the token respectively. (b) The compound words are separated by a single '\#' token. (c) The tokens tokenized by BPE, with a leading space have a 'Ġ' symbol. (d) The BPE splits and the stems from MorphTable have no special symbol in them. Classification of tokens with surface forms of the first three types is straightforward. For the tokens that have no special symbol, we have a heuristically driven algorithm that marks them either 'morph/stem' or 'bpe'.

\begin{figure}
    \includegraphics[width=0.4\textwidth]{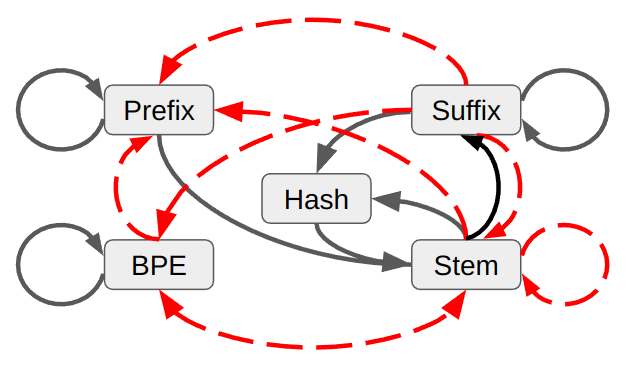}
    \centering
    \caption{Detokenization mechanism from morphemes to English words. Black lines show word continuation; red dashed lines show word boundary, and missing connections imply invalid transitions. Hash denotes compound words. Stem$>$Stem is a special case}
    \label{fig:detokenizer}
\end{figure}

\subsection{Reverse MorphTable}
\label{rev_morph}
Once all tokens are classified as above, the 'bpe' tokens are combined together following the standard BPE algorithm, which essentially involves concatenating them together and using byte pair decoding. However, for the tokens marked 'morph', the procedure is more involved. First we need to find tokens that are morpheme constituents of the same word (i.e find word boundaries) and then use a reverse MorphTable to find those words. Finding word boundaries is further complicated by various cases like compound words, multiple affixes etc. To combine various cases of these surface forms, I developed a heuristic algorithm (Figure \ref{fig:detokenizer}) that gives us word continuation and word boundaries between different tokens. This algorithm defines sequence of surface forms that would form a valid segmentation for a word by looking at consecutive token labels from Section \ref{classify_tokens}. Once the word boundaries are found, the reverse-MorphTable is then used to convert this segmentation to an English word. 

\subsection{Illustrative Example}

Let's assume, a model trained on MorphPiece outputs the tokens shown in Figure \ref{fig:detoken_example}. In the first step, the tokens will get classified as : ['bpe', 'bpe', 'prefix', 'stem', 'suffix', 'stem', 'suffix'] with additional label of 'morph' on all tokens except those marked 'bpe'. Since merging of tokens labelled 'bpe' is straightforward, we focus on those marked 'morph'. Now we follow the arrows from Figure \ref{fig:detokenizer}; with the solid black lines showing word continuation and red dashed lines showing word boundaries. From here we get word boundaries as : ['in\#','vestigate','ing'] and ['diligent','\#ly']. Finally we look up the words in a reverse-MorphTable to get 'investigating' and 'diligently'.

\section{Discussion}
\label{discussion}

Generally token level perplexity is not comparable across different vocabularies \cite{cotterell-etal-2018-languages}. In this work I contend that token level perplexity is simply an indication of how well the model/tokenizer performs on the next-token-prediction language modeling task. By splitting the tokens in a more aligned manner, this task gets easier which results in lower token level perplexity. Moreover, not only does this improved performance translate into improvement in a wide variety of other NLP tasks, but it also reflects in the LAMBADA task, which is pretty close to language modeling task. 

Another way to look at it is that MorphPiece shifts the probability mass on tokens in a way that is potentially more suitable for language modeling task. With BPE, the model does all the heavy lifting of finding the probability distribution. By providing a less ‘noisy’ input space (i.e. aligned canonical morpheme segmentations) the tokenizer plays a more active role in finding better probability distributions.

Finally, MorphPiece not only splits words into morphemes; but by using a manually curated dictionary, we get canonical segmentation ('bat','ing') instead of surface-form segmentation ('bat','ting' or 'batt','ing') \cite{cotterell, batsuren-etal-2022-sigmorphon}. This, along with the alignment of subwords into affixes, is arguably the underlying reason for better MorphPiece performance. On the other hand, a model trained on BPE has to tackle with all sort of noisy segmentations. 

\section*{Limitations, Caveats and Way Forward}
There are a couple of caveats/limitations of MorphPiece in its current state, however all of them are rather straightforward to fix in derivative works, and except for the last one, not likely to influence the results of this paper much.

\begin{itemize}
    \item First, the MorphTable was constructed from MorphyNet. During the course of research, it was observed that it does not cover all lexical families. 
    \item Second, for each language, we have to separately construct the MorphTable and detokenization automata (Figure \ref{fig:detokenizer}) to find word boundaries and continuations.
    \item Third, for English, this tokenization produces 17\% more tokens compared to BPE. However, reducing these extra tokens by fusing morphemes can be taken up in a derivative work.
    \item While the detokenization algorithm works for standard English language, it might fail for missepelled words where word boundaries are ill defined.
    \item For multilingual tokenizers, there will be an extra step to first classify which language the tokens belong to and then use the respective detokenization scheme. Since consecutive output sequences are usually monolingual, it is unlikely to add significant inference costs.
    \item Lastly, MorphPiece needs to be trained on a larger model to see if the gains scale with model size.
\end{itemize}

\section*{Conclusion}

In conclusion, I have presented a linguistically motivated tokenization scheme that outperforms models trained on BPE on a wide variety of tasks. I hope that this paradigm of using linguistic inductive bias will lay the foundations of a new generation of models that move away from purely statistical language representation.

\section*{Ethical/Societal Impact}
This paper presents work whose goal is to advance the field of Large Language Models with a focus on linguistic composition of languages. There are many potential societal consequences of LLMs in general, none which we feel must be specifically highlighted here. However, the linguistic representation of text is likely to have a positive impact on development of LLMs, especially for low resource morphologically rich languages.

\bibliography{morphpiece}
\bibliographystyle{icml2024}

\newpage
\appendix
\onecolumn
\section{Appendices}

\begin{table}[H]
\centering
\begin{tabular}{@{}cccc@{}}
\toprule
\textbf{Dataset} & \textbf{GPT-2} & \textbf{MorphGPT} & \textbf{Diff (\%)} \\ \midrule
\textbf{AmazonCounterfactual} & 0.729 & 0.705 & -3.335 \\
\textbf{AmazonPolarity} & 0.597 & 0.655 & 9.803 \\
\textbf{AmazonReviews} & 0.270 & 0.310 & 15.079 \\
\textbf{Banking77} & 0.331 & 0.476 & 44.018 \\
\textbf{Emotion} & 0.259 & 0.292 & 12.596 \\
\textbf{Imdb} & 0.599 & 0.675 & 12.595 \\
\textbf{MTOPDomain} & 0.544 & 0.689 & 26.571 \\
\textbf{MTOPIntent} & 0.352 & 0.557 & 58.024 \\
\textbf{MassiveIntent} & 0.335 & 0.432 & 29.111 \\
\textbf{MassiveScenario} & 0.396 & 0.478 & 20.583 \\
\textbf{ToxicConversations} & 0.630 & 0.674 & 6.927 \\
\textbf{TweetSentimentExtraction} & 0.469 & 0.496 & 5.833 \\ \bottomrule
\end{tabular}
\caption{Results of Classification Task in MTEB. Scores are Accuracy}
\label{tab:mteb_classification}
\end{table}

\begin{table}[H]
\centering
\begin{tabular}{ccccc}
\toprule
\textbf{Dataset} & \textbf{Metric} & \textbf{GPT-2} & \textbf{MorphGPT} & \textbf{Diff (\%)} \\ \midrule
\multirow{2}{*}{\textbf{ArxivClusteringP2P}} & \textbf{v\_measure} & 0.144 & \textbf{0.297} & 105.952 \\
 & \textbf{v\_measure\_std} & 0.171 & \textbf{0.152} & -11.370 \\
\multirow{2}{*}{\textbf{ArxivClusteringS2S}} & \textbf{v\_measure} & 0.105 & \textbf{0.156} & 48.727 \\
 & \textbf{v\_measure\_std} & 0.171 & \textbf{0.165} & -3.552 \\
\multirow{2}{*}{\textbf{BiorxivClusteringP2P}} & \textbf{v\_measure} & 0.106 & \textbf{0.255} & 140.346 \\
 & \textbf{v\_measure\_std} & 0.005 & \textbf{0.009} & 75.795 \\
\multirow{2}{*}{\textbf{BiorxivClusteringS2S}} & \textbf{v\_measure} & 0.064 & \textbf{0.128} & 98.340 \\
 & \textbf{v\_measure\_std} & 0.005 & 0.006 & 8.028 \\
\multirow{2}{*}{\textbf{MedrxivClusteringP2P}} & \textbf{v\_measure} & 0.145 & \textbf{0.244} & 67.598 \\
 & \textbf{v\_measure\_std} & 0.021 & \textbf{0.017} & -18.270 \\
\multirow{2}{*}{\textbf{MedrxivClusteringS2S}} & \textbf{v\_measure} & 0.144 & \textbf{0.178} & 23.849 \\
 & \textbf{v\_measure\_std} & 0.022 & \textbf{0.019} & -16.844 \\
\multirow{2}{*}{\textbf{RedditClustering}} & \textbf{v\_measure} & 0.074 & \textbf{0.184} & 148.378 \\
 & \textbf{v\_measure\_std} & \textbf{0.022} & 0.026 & 16.028 \\
\multirow{2}{*}{\textbf{RedditClusteringP2P}} & \textbf{v\_measure} & 0.131 & \textbf{0.385} & 193.846 \\
 & \textbf{v\_measure\_std} & \textbf{0.063} & 0.110 & 74.594 \\
\multirow{2}{*}{\textbf{StackExchangeClustering}} & \textbf{v\_measure} & 0.109 & \textbf{0.319} & 191.634 \\
 & \textbf{v\_measure\_std} & \textbf{0.036} & 0.041 & 13.673 \\
\multirow{2}{*}{\textbf{StackExchangeClusteringP2P}} & \textbf{v\_measure} & 0.269 & \textbf{0.306} & 13.708 \\
 & \textbf{v\_measure\_std} & 0.024 & \textbf{0.020} & -14.964 \\
\multirow{2}{*}{\textbf{TwentyNewsgroupsClustering}} & \textbf{v\_measure} & 0.074 & \textbf{0.180} & 143.054 \\
 & \textbf{v\_measure\_std} & 0.021 & \textbf{0.017} & -18.373 \\ \bottomrule 
\end{tabular}
\caption{Results on Clustering Task on MTEB Benchmark}
\label{tab:mteb_clustering}
\end{table}

\begin{table}
\centering
\begin{tabular}{ccccccc}
\toprule
\textbf{Metric} & \textbf{GPT-2} & \textbf{MorphGPT} & \multicolumn{1}{c|}{\textbf{Diff (\%)}} & \textbf{GPT-2} & \textbf{MorphGPT} & \textbf{Diff (\%)} \\ \cmidrule(l){2-7} 
\textbf{} & \multicolumn{3}{c|}{\textbf{Average Precision}} & \multicolumn{3}{c}{\textbf{F1 Score}} \\ \midrule 
\textbf{cos\_sim} & 0.274 & \textbf{0.439} & 191.125 & 0.286 & \textbf{0.446} & 130.235 \\
\textbf{dot} & 0.155 & \textbf{0.332} & 190.519 & 0.265 & \textbf{0.360} & 119.258 \\
\textbf{euclidean} & 0.264 & \textbf{0.422} & 251.981 & 0.291 & \textbf{0.431} & 174.334 \\
\textbf{manhattan} & 0.449 & \textbf{0.508} & 18.356 & 0.443 & \textbf{0.497} & 14.668 \\ 
\textbf{max} & 0.449 & \textbf{0.508} & 18.356 & 0.443 & \textbf{0.497} & 14.668 \\ \bottomrule
\end{tabular}
\caption{Results on Pair Classification Task on MTEB}
\label{tab:mteb_pair}
\end{table}

\begin{table}
\centering
\begin{tabular}{@{}cccc@{}}
\toprule
\textbf{Dataset} & \textbf{GPT-2} & \textbf{MorphGPT} & \textbf{Diff (\%)} \\ \midrule
\textbf{ArguAna} & 0.122 & \textbf{0.257} & 111.39 \\
\textbf{CQADupstackAndroidRetrieval} & 0.016 & \textbf{0.040} & 144.38 \\
\textbf{CQADupstackEnglishRetrieval} & 0.005 & \textbf{0.026} & 385.57 \\
\textbf{CQADupstackGamingRetrieval} & 0.009 & \textbf{0.031} & 246.46 \\
\textbf{CQADupstackGisRetrieval} & 0.010 & \textbf{0.015} & 61.55 \\
\textbf{CQADupstackMathematicaRetrieval} & 0.003 & \textbf{0.006} & 100.93 \\
\textbf{CQADupstackPhysicsRetrieval} & 0.010 & \textbf{0.034} & 221.47 \\
\textbf{CQADupstackProgrammersRetrieval} & 0.006 & \textbf{0.016} & 183.77 \\
\textbf{CQADupstackStatsRetrieval} & 0.001 & \textbf{0.015} & 1584.56 \\
\textbf{CQADupstackTexRetrieval} & 0.005 & \textbf{0.010} & 107.33 \\
\textbf{CQADupstackUnixRetrieval} & 0.006 & \textbf{0.024} & 319.85 \\
\textbf{CQADupstackWebmastersRetrieval} & 0.010 & \textbf{0.020} & 100.41 \\
\textbf{CQADupstackWordpressRetrieval} & 0.000 & \textbf{0.020} & Inf \\
\textbf{SCIDOCS} & 0.002 & \textbf{0.005} & 123.78 \\
\textbf{Touche2020} & 0.008 & \textbf{0.016} & 104.11 \\
\textbf{TRECCOVID} & 0.048 & \textbf{0.093} & 93.40 \\ \bottomrule
\end{tabular}
\caption{Results on MTEB Retrieval Task at NDCG@10}
\label{tab:mteb_retrieval}
\end{table}

\begin{table}
\centering
\begin{tabular}{@{}cccc|rrr@{}}
\toprule
\multicolumn{1}{c}{\multirow{2}{*}{\textbf{STS Dataset}}} & \textbf{GPT-2} & \textbf{MorphGPT} & \textbf{Diff (\%)} & \multicolumn{1}{c}{\textbf{GPT-2}} & \multicolumn{1}{c}{\textbf{MorphGPT}} & \multicolumn{1}{c}{\textbf{Diff (\%)}} \\ \cmidrule(l){2-7} 
\multicolumn{1}{c}{} & \multicolumn{3}{c|}{\textbf{Spearman Correlation}} & \multicolumn{3}{c}{\textbf{Pearson Correlation}} \\ \midrule 
\textbf{BIOSSES} & 0.324 & \textbf{0.456} & 40.893 & 0.202 & \textbf{0.349} & 72.538 \\
\textbf{SICK-R} & 0.438 & \textbf{0.488} & 11.238 & 0.360 & \textbf{0.516} & 43.103 \\
\textbf{STS12} & 0.258 & \textbf{0.399} & 54.415 & 0.122 & \textbf{0.360} & 196.388 \\
\textbf{STS13} & 0.289 & \textbf{0.521} & 80.357 & 0.217 & \textbf{0.488} & 125.117 \\
\textbf{STS14} & 0.262 & \textbf{0.412} & 57.370 & 0.197 & \textbf{0.414} & 109.732 \\
\textbf{STS15} & 0.347 & \textbf{0.557} & 60.314 & 0.270 & \textbf{0.536} & 98.758 \\
\textbf{STS16} & 0.357 & \textbf{0.508} & 42.436 & 0.208 & \textbf{0.473} & 127.901 \\
\textbf{STSBenchmark} & 0.259 & \textbf{0.403} & 55.388 & 0.165 & \textbf{0.386} & 133.699 \\ \bottomrule
\end{tabular}
\caption{Results on STS task of MTEB. The metric is cosine similarity}
\label{tab:mteb_sts}
\end{table}

\begin{table}
\centering
\begin{tabular}{@{}rrrrr@{}}
\toprule
\textbf{Correlation} & \textbf{Metric} & \textbf{GPT-2} & \textbf{MorphGPT} & \textbf{Diff (\%)} \\ \midrule
\multirow{2}{*}{\textbf{Pearson}} & \textbf{cos\_sim} & 0.289094 & \textbf{0.294273} & 1.791419 \\
 & \textbf{dot} & 0.206547 & \textbf{0.258185} & 25.000629 \\
\multirow{2}{*}{\textbf{Spearman}} & \textbf{cos\_sim} & 0.286151 & \textbf{0.295015} & 3.097671 \\
 & \textbf{dot} & 0.224946 & \textbf{0.274230} & 21.908899 \\ \bottomrule
\end{tabular}
\caption{Evaluation on Summarization Task of MTEB}
\label{tab:mteb_summary}
\end{table}

\begin{table}
\centering
\begin{tabular}{@{}ccccc@{}}
\toprule
\textbf{Dataset Name} & \textbf{Metric} & \textbf{GPT-2} & \textbf{MorphGPT} & \textbf{Diff (\%)} \\ \midrule
\multirow{2}{*}{\textbf{AskUbuntuDupQuestions}} & \textbf{map} & 0.404 & \textbf{0.454} & 12.501 \\
 & \textbf{mrr} & 0.516 & \textbf{0.571} & 10.666 \\
\multirow{2}{*}{\textbf{MindSmallReranking}} & \textbf{map} & \textbf{0.269} & 0.252 & -6.479 \\
 & \textbf{mrr} & \textbf{0.274} & 0.251 & -8.437 \\
\multirow{2}{*}{\textbf{SciDocsRR}} & \textbf{map} & 0.371 & \textbf{0.436} & 17.643 \\
 & \textbf{mrr} & 0.579 & \textbf{0.672} & 16.092 \\
\multirow{2}{*}{\textbf{StackOverflowDupQuestions}} & \textbf{map} & 0.245 & \textbf{0.302} & 23.233 \\
 & \textbf{mrr} & 0.231 & \textbf{0.294} & 27.501 \\ \bottomrule
\end{tabular}
\caption{Results on ReRanking Task (MTEB)}
\label{tab:mteb_reranking}
\end{table}

\begin{table}
\begin{tabular}{@{}ccccccccccccc@{}}
\toprule
\textbf{size} & \multicolumn{6}{c}{\textbf{small}} & \multicolumn{6}{c}{\textbf{large}} \\ \cmidrule(l){2-13} 
\textbf{} & \multicolumn{3}{c}{\textbf{f1\_dev}} & \multicolumn{3}{c}{\textbf{f1\_test}} & \multicolumn{3}{c}{\textbf{f1\_dev}} & \multicolumn{3}{c}{\textbf{f1\_test}} \\ \cmidrule(l){2-13} 
\textbf{domain} & \textbf{cs} & \textbf{maths} & \textbf{physics} & \textbf{cs} & \textbf{maths} & \multicolumn{1}{c|}{\textbf{physics}} & \textbf{cs} & \textbf{maths} & \textbf{physics} & \textbf{cs} & \textbf{maths} & \textbf{physics} \\ \midrule
\textbf{GPT-2} & 0.355 & 0.256 & 0.368 & 0.334 & 0.283 & \multicolumn{1}{c|}{0.423} & 0.490 & 0.506 & 0.586 & 0.468 & 0.498 & 0.564 \\
\textbf{+FLOTA} & 0.340 & 0.310 & 0.400 & 0.358 & 0.309 & \multicolumn{1}{c|}{0.446} & 0.511 & 0.548 & 0.602 & 0.496 & 0.529 & 0.585 \\
\textbf{MorphGPT} & \textbf{0.526} & \textbf{0.478} & \textbf{0.529} & \textbf{0.514} & \textbf{0.475} & \multicolumn{1}{c|}{\textbf{0.580}} & \textbf{0.669} & \textbf{0.647} & \textbf{0.683} & \textbf{0.644} & \textbf{0.651} & \textbf{0.662} \\ \bottomrule
\end{tabular}
\caption{Detailed results on FLOTA (without noise)}
\label{tab:flota_all}
\end{table}

\begin{table}
\begin{tabular}{@{}ccccccccccccc@{}}
\toprule
\textbf{size} & \multicolumn{6}{c}{\textbf{small}} & \multicolumn{6}{c}{\textbf{large}} \\ \cmidrule(l){2-13} 
\textbf{} & \multicolumn{3}{c}{\textbf{f1\_dev}} & \multicolumn{3}{c}{\textbf{f1\_test}} & \multicolumn{3}{c}{\textbf{f1\_dev}} & \multicolumn{3}{c}{\textbf{f1\_test}} \\ \cmidrule(l){2-13} 
\multicolumn{1}{c}{\textbf{domain}} & \textbf{cs} & \textbf{maths} & \textbf{physics} & \textbf{cs} & \textbf{maths} & \multicolumn{1}{c|}{\textbf{physics}} & \textbf{cs} & \textbf{maths} & \textbf{physics} & \textbf{cs} & \textbf{maths} & \textbf{physics} \\ \midrule
\multicolumn{1}{c}{\textbf{GPT-2}} & 0.258 & 0.199 & 0.277 & 0.270 & 0.218 & \multicolumn{1}{c|}{0.343} & 0.413 & 0.400 & 0.440 & 0.407 & 0.390 & 0.421 \\
\multicolumn{1}{c}{\textbf{+FLOTA}} & 0.232 & 0.229 & 0.290 & 0.283 & 0.192 & \multicolumn{1}{c|}{0.322} & 0.469 & 0.408 & 0.503 & 0.451 & 0.400 & 0.482 \\
\multicolumn{1}{c}{\textbf{MorphGPT}} & \textbf{0.478} & \textbf{0.436} & \textbf{0.471} & \textbf{0.463} & \textbf{0.412} & \multicolumn{1}{c|}{\textbf{0.514}} & \textbf{0.596} & \textbf{0.565} & \textbf{0.598} & \textbf{0.567} & \textbf{0.557} & \textbf{0.580} \\ \bottomrule
\end{tabular}
\caption{Detailed results on FLOTA (with noise)}
\label{tab:flota_noise_all}
\end{table}

\begin{figure}
    \centering
    \includegraphics[width=1.1\textwidth]{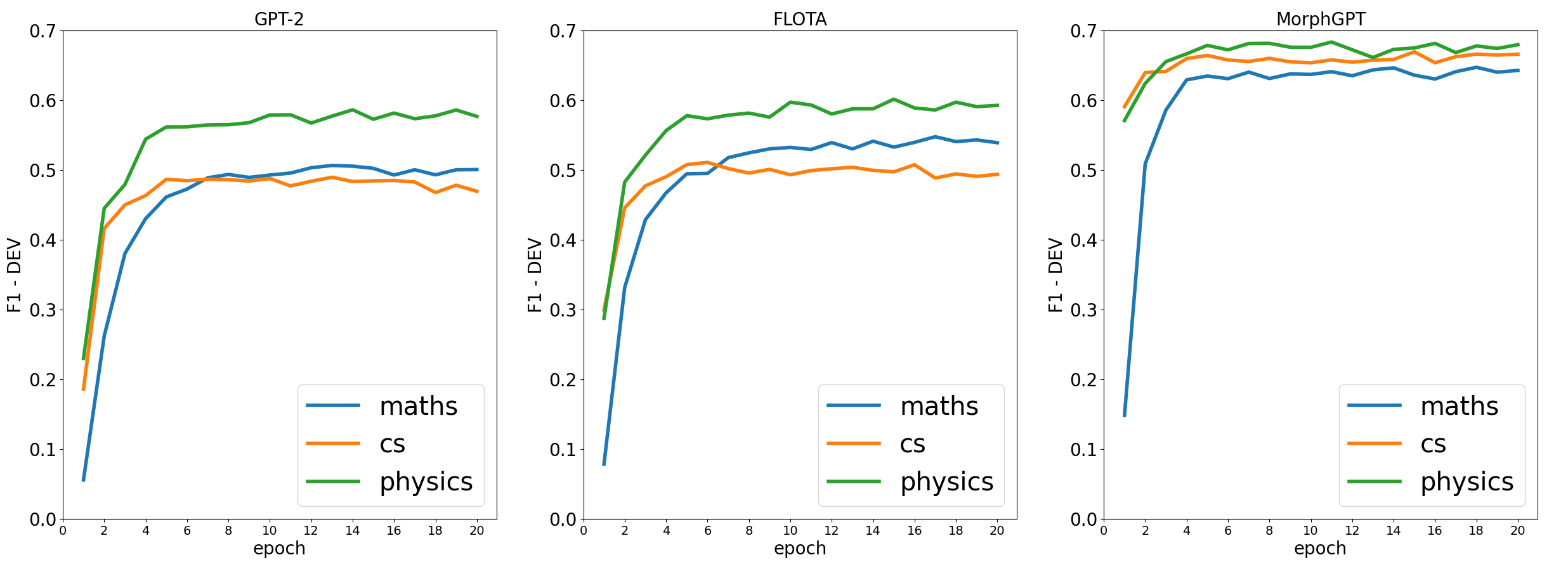}
    \caption{Training performance of GPT-2, GPT-2+FLOTA and MorphGPT on ArXiv-Large datasets.}
    \label{fig:flota}
\end{figure}

\end{document}